\documentclass[runningheads]{llncs}
\usepackage[T1]{fontenc}
\usepackage{graphicx}
\usepackage{xcolor}
\usepackage{multirow}
\begin{document}
\title{ICDAR 2023 Competition on Hierarchical Text Detection and Recognition}
\author{Shangbang Long \and
Siyang Qin \and
Dmitry Panteleev \and
Alessandro Bissacco \and
Yasuhisa Fujii \and
Michalis Raptis
}
\authorrunning{S. Long et al.}
\institute{Google Research \\
\email{\{longshangbang,qinb,dpantele,bissacco,yasuhisaf,mraptis\}@google.com}}
\maketitle              % typeset the header of the contribution
\begin{abstract}
We organize a competition on hierarchical text detection and recognition.
The competition is aimed to promote research into deep learning models and systems that can jointly perform text detection and recognition and geometric layout analysis.
We present details of the proposed competition organization, including tasks, datasets, evaluations, and schedule.
During the competition
period (from January 2nd 2023 to April 1st 2023), at least $50$ submissions from more than $20$ teams were made in the $2$ proposed tasks.
Considering the number of teams and submissions, we conclude that the HierText competition has been successfully held.
In this report, we will also present the competition results and insights from them.


\end{abstract}

%
%
%
% \begin{table}
% \caption{Table captions should be placed above the
% tables.}\label{tab1}
% \begin{tabular}{|l|l|l|}
% \hline
% Heading level &  Example & Font size and style\\
% \hline
% Title (centered) &  {\Large\bfseries Lecture Notes} & 14 point, bold\\
% 1st-level heading &  {\large\bfseries 1 Introduction} & 12 point, bold\\
% 2nd-level heading & {\bfseries 2.1 Printing Area} & 10 point, bold\\
% 3rd-level heading & {\bfseries Run-in Heading in Bold.} Text follows & 10 point, bold\\
% 4th-level heading & {\itshape Lowest Level Heading.} Text follows & 10 point, italic\\
% \hline
% \end{tabular}
% \end{table}

% \begin{figure}
% \includegraphics[width=\textwidth]{fig1.eps}
% \caption{A figure caption is always placed below the illustration.
% Please note that short captions are centered, while long ones are
% justified by the macro package automatically.} \label{fig1}
% \end{figure}

\section{Introduction}
Text detection and recognition systems \cite{survey_2021} and geometric layout analysis techniques \cite{Lee_2019,yang_2017} have long been developed separately as independent tasks.
Research on text detection and recognition \cite{ronen_2022,long_2018,qin_2019,kittenplon_2022} has mainly focused on the domain of natural images and aimed at single level text spotting (mostly, word-level).
Conversely, research on geometric layout analysis \cite{Lee_2019,yang_2017,liu_2022,wang_2022}, which is targeted at parsing text paragraphs and forming text clusters, has assumed document images as input and taken OCR results as fixed and given by independent systems.
The synergy between the two tasks remains largely under-explored.

Recently, the Unified Detector work by Long et al. \cite{long_2022} shows that the unification of line-level detection of text and geometric layout analysis benefits both tasks significantly. 
StructuralLM \cite{li_2021} and LayoutLMv3 \cite{huang_2022} show that text line grouping signals are beneficial to the downstream task of document understanding and are superior to word-level bounding box signals.
These initial studies demonstrate that the unification of OCR and layout analysis, which we term as \textit{Hierarchical Text Detection and Recognition (HTDR)}, can be mutually beneficial to OCR, layout analysis, and downstream tasks.

\begin{figure}
\includegraphics[width=\textwidth]{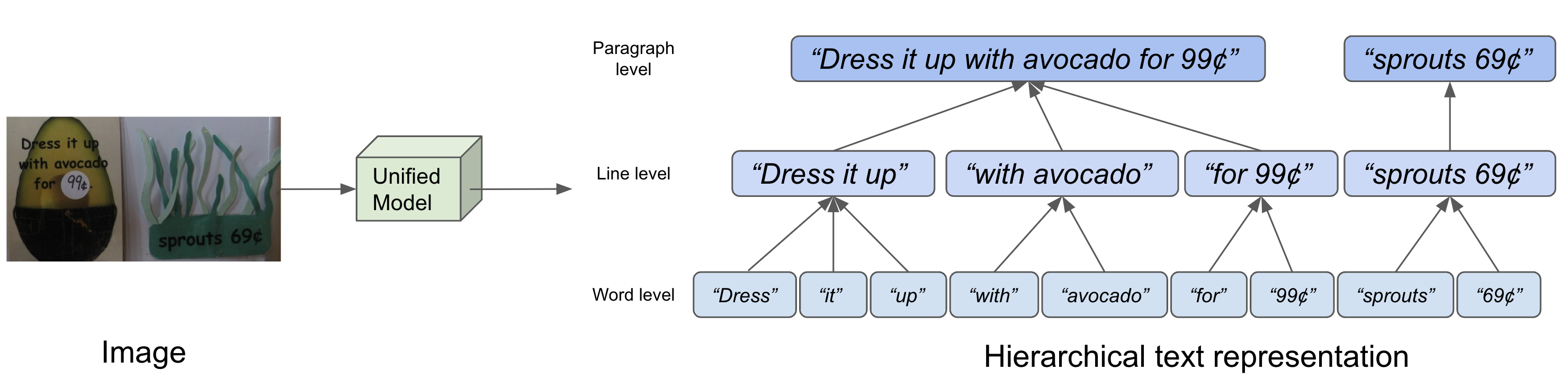}
\caption{Illustration for the proposed unified task: \textbf{Hierarchical Text Detection and Recognition (HTDR)}. 
Given an input image, the unified model is expected to produce a hierarchical text representation, which resembles the form of a forest. 
Each tree in the forest represents one paragraph and has three layers, representing the clustering of words into lines and then paragraphs.
} \label{fig-overview}
\end{figure}

Given the promising potential benefits, we propose the \textbf{ICDAR 2023 Competition on Hierarchical Text Detection and Recognition}.
In this competition, candidate systems are expected to perform the unified task of text detection and recognition and geometric layout analysis.
Specifically, we define the unified task as producing a hierarchical text representation, including word-level bounding boxes and text transcriptions, as well as line-level and paragraph-level clustering of these word-level text entities. 
We defer the rigorous definitions of word / line / paragraph later to the dataset section. 
Fig. \ref{fig-overview} illustrates our notion of the unified task.

% TODO: Change the `Image` to a more proper one.

We believe this competition will have profound and long-term impact on the whole image-based text understanding field by unifying the efforts of text detection and recognition and geometric layout analysis, and furthering providing new signals for downstream tasks. 

The competition started on January 2nd 2023, received more than $50$ submissions in $2$ tasks in total, and closed on April 1st 2023. 
This report provides details into the motivation, preparation, and results of the competition.
We believe the success of this competition greatly promotes the development of this research field.
Furthermore, the dataset except the test set annotation and evaluation script are made publicly available.
The competition website\footnote{\url{https://rrc.cvc.uab.es/?ch=18}} remains open to submission and provides evaluation on the test set.

\section{Competition Protocols}

\subsection{Dataset}
The competition is based on the HierText dataset \cite{long_2022}.
Images in HierText are collected from the Open Images v6 dataset \cite{kuz_2020}, by first applying the \textit{Google Cloud Platform (GCP) Text Detection API}\footnote{\url{https://cloud.google.com/vision/docs/ocr}} and then filtering out inappropriate images, for example those with too few text or non-English text.
In total, $11639$ images are obtained.
In this competition, we follow the original split of $8281/1724/1634$ for \textit{train}, \textit{validation}, \textit{test} sets.
Images and annotations of the train and validation set are released publicly. 
The test set annotation is kept private and will remain so even after the end of the competition.

% singh_2021,krylov_2021
As noted in the original paper \cite{long_2022}, we check the cross-dataset overlap rates with the two other OCR datasets that are based on Open Images.
We find that $1.5\%$ of the $11639$ images we have are also in TextOCR \cite{singh_2021} and $3.6\%$ in Intel OCR \cite{krylov_2021}.
Our splits ensure that our training images are not in the validation or test set of Text OCR and Intel OCR, and vice versa.

\begin{figure}
\centering
\includegraphics[width=1.0\textwidth]{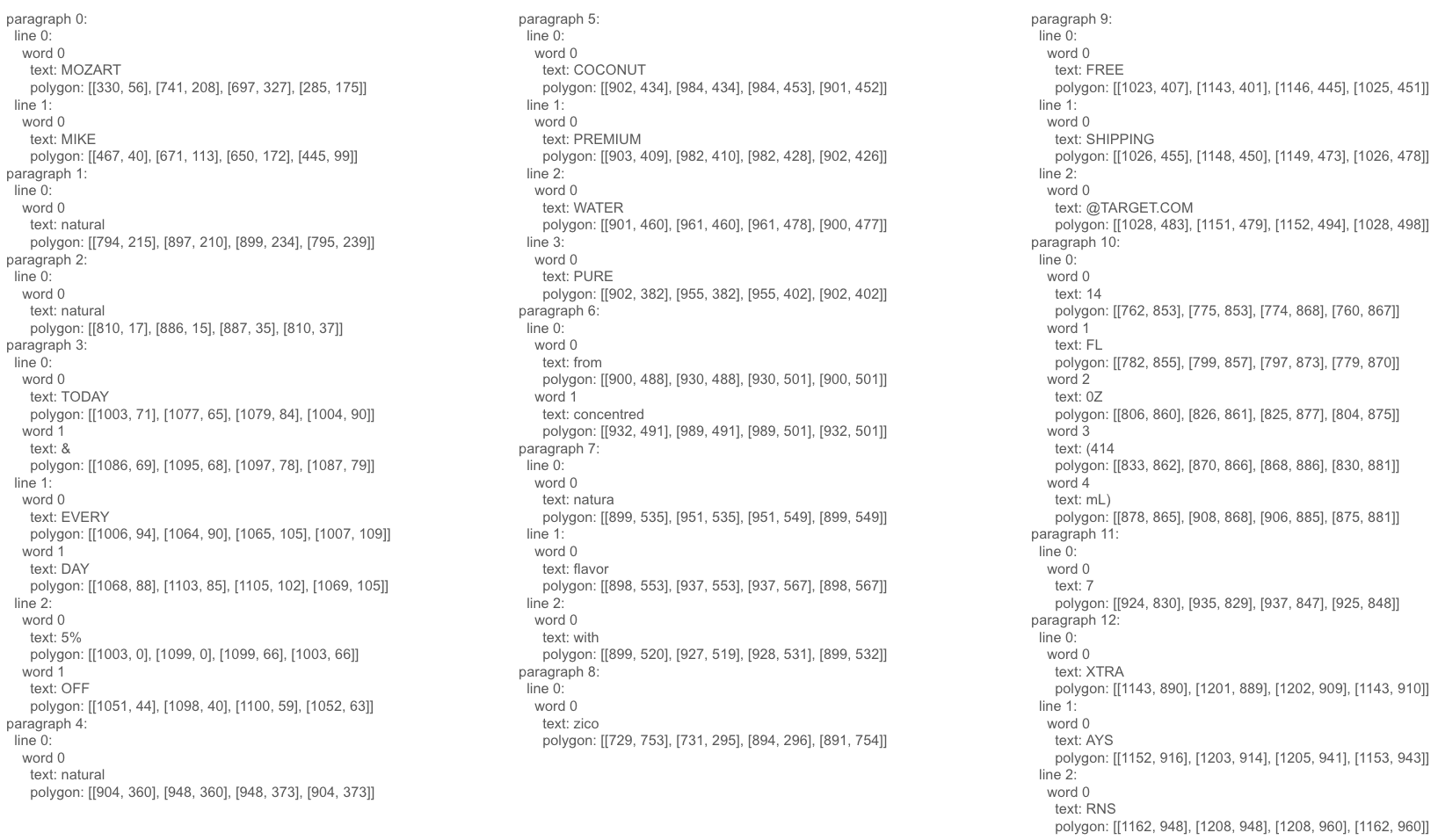}
\caption{
Example of hierarchical annotation format of the dataset.
}
\label{fig-ann}
\end{figure}

\begin{figure}

Sample 1

\includegraphics[width=\textwidth]{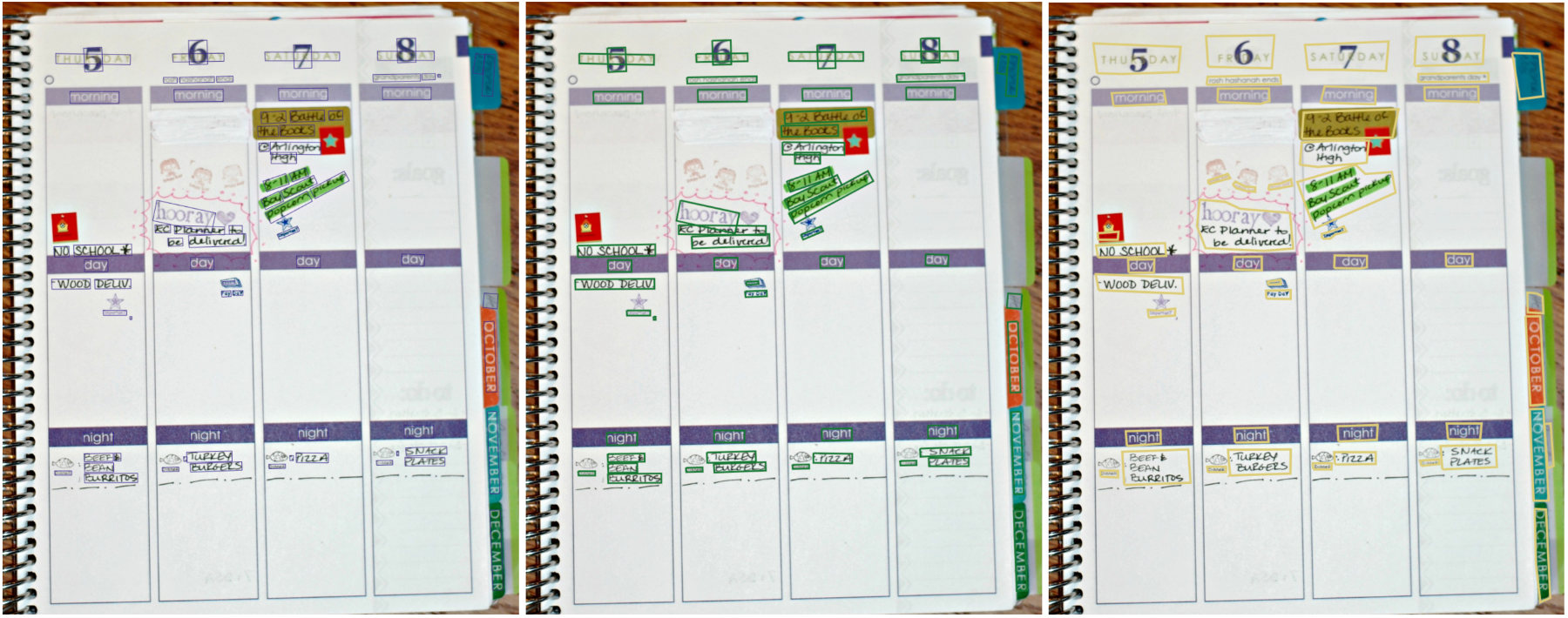}

Sample 2

\includegraphics[width=\textwidth]{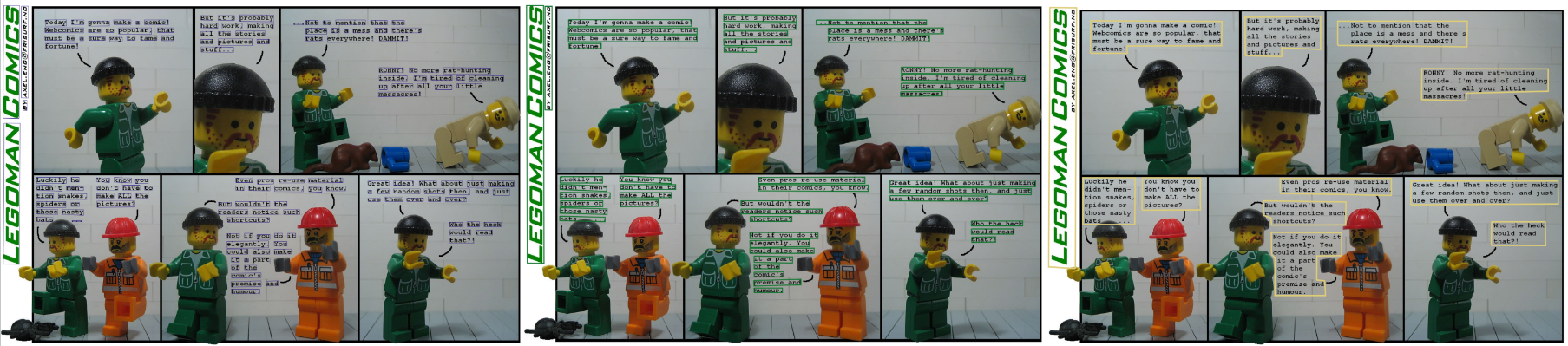}

Sample 3

\includegraphics[width=\textwidth]{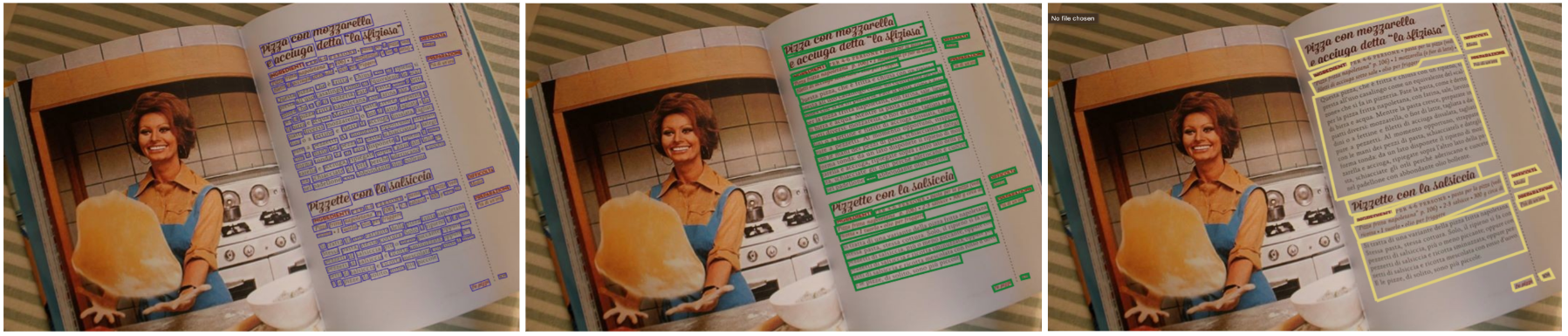}

Sample 4

\includegraphics[width=\textwidth]{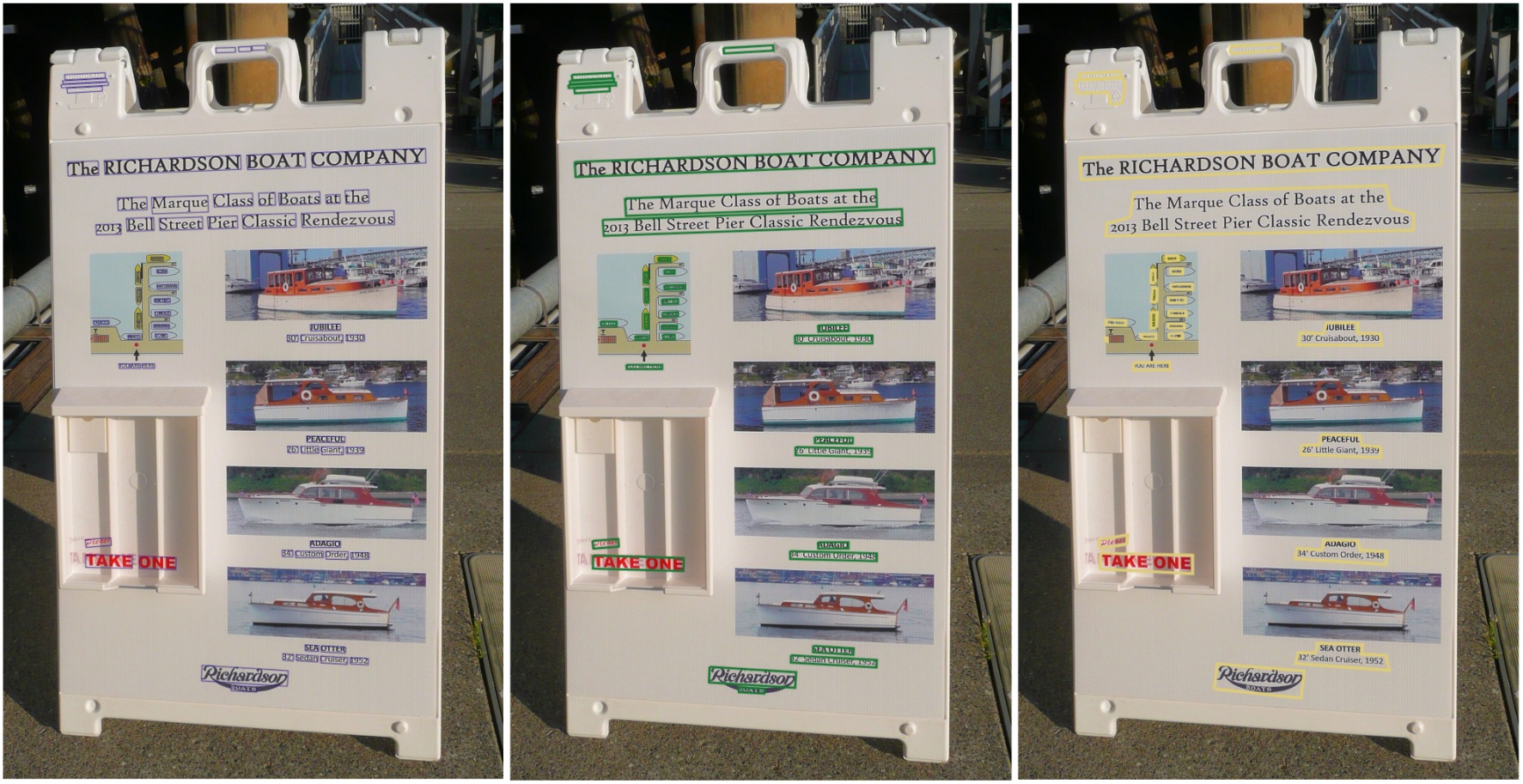}

\caption{
Illustration for the hierarchical annotation of text in images.
From \textbf{left} to \textbf{right}: \textbf{word}, \textbf{line}, \textbf{paragraph} level annotations.
Words (\textcolor{blue}{blue}) are annotated with polygons.
Lines (\textcolor{teal}{green}) and paragraphs (\textcolor{yellow}{yellow}) are annotated as hierarchical clusters and visualized as polygons.
Images are taken from the train split.
}
\label{fig-annotation}
\end{figure}

The images are annotated in a hierarchical way of \textit{word}-to-\textit{line}-to-\textit{paragraph}, as shown in Fig. \ref{fig-ann}. 
\textit{Words} are defined as a sequence of textual characters not interrupted by \textit{spaces}.
\textit{Lines} are then defined as \textit{space}-separated clusters of \textit{words} that are logically connected and aligned in spatial proximity.
Finally, \textit{paragraphs} are composed of  \textit{lines} that belong to the same semantic topic and are geometrically coherent.
Fig. \ref{fig-annotation} illustrates some annotated samples.
Words are annotated with polygons, with $4$ vertices for straight text and more for curved text depending on the shape.
Then, words are transcribed regardless of the scripts and languages, as long as they are legible.
Note that we do not limit the character sets, so the annotation could contain case-sensitive characters, digits, punctuation, as well as non-Latin characters such as Cyrillic and Greek.
After word-level annotation, we group words into lines and then group lines into paragraphs.
In this way, we obtain a hierarchical annotation that resembles a forest structure of the text in an image.

\subsection{Tasks}

Our challenge consists of $2$ competition tracks, \textbf{Hierarchical Text Detection} and \textbf{Word-Level End-to-End Text Detection and Recognition}. 
In the future, we plan to merge them into a single unified Hierarchical Text Spotting task that requires participants to give a unified representation of text with layout.

\subsubsection{Task 1: Hierarchical Text Detection}
This task itself is formulated as a combination of $3$ tasks: word detection, text line detection, and paragraph detection, where lines and paragraphs are represented as clusters of words hierarchically.

In this task, participants are provided with images and expected to produce the hierarchical text detection results.
Specifically, the results are composed of \textbf{word-level bounding polygons} and \textbf{line and paragraph clusters} on top of words.
The clusters are represented as forests, as in Fig. \ref{fig-overview}, where each paragraph is a tree and words are leaves.
For this task, participants do not need to provide text recognition results. 

\begin{figure}
\centering
\includegraphics[width=1.0\textwidth]{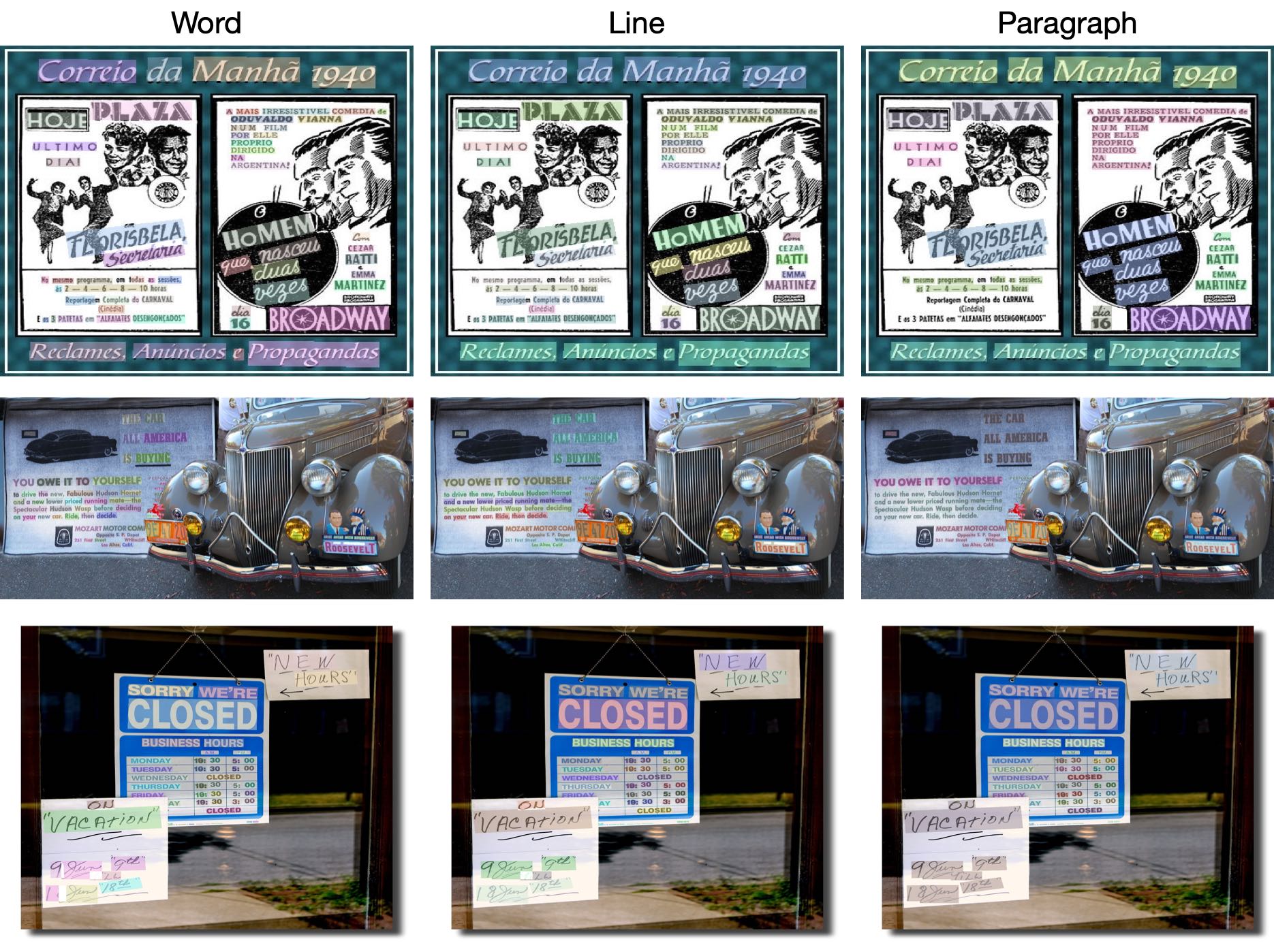}
\caption{Illustration of how hierarchical text detection can be evaluated as $3$ instance segmentation sub-tasks.
The coloring of each column indicates the instance segmentation for each sub-task.
} \label{fig-evaluation}
\end{figure}

As illustrated in Fig. \ref{fig-evaluation}, we evaluate this task as $3$ instance segmentation sub-tasks for word, line, and paragraph respectively.
For word level, each word is one instance.
For line level, we take the union of each line's children words as one instance.
For paragraph level, we aggregate each paragraph's children lines, and take that as one instance. 
With this formulation, all the 3 sub-tasks will be evaluated with the PQ metric \cite{kirillov_2019} designed for instance segmentation, as specified in \cite{long_2022}:

\begin{equation}
  PQ = \frac{\sum_{(p, g)\in TP}IoU(p, g)}{|TP| + \frac{1}{2}|FP| + \frac{1}{2}|FN|}
  \label{eq:pq-def}
\end{equation}

\noindent where $TP,~FP,~FN$ represent  true positives, false positives, and false negatives respectively. 
We use an IoU threshold of $0.5$ to count true positives. 
Note that the PQ metric is mathematically equal to the product of the \textit{Tightness} score, which is defined as the average IoU scores of all TP pairs, and the \textit{F1}, score which is commonly used in previous OCR benchmarks. 
Previous OCR evaluation protocols only report F1 scores which do not fully reflect the detection quality.
We argue that tightness is very important in evaluating hierarchical detection. 
It gives an accurate measurement of how well detections match ground-truths. 
For words, a detection needs to enclose all its characters and not overlap with other words, so that the recognition can be correct.
The tightness score can penalize missing characters and oversized boxes.
For lines and paragraphs, they are represented as clusters of words, and are evaluated as unions of masks.
Wrong clustering of words can also be reflected in the IoU scores for lines and paragraphs.
In this way, using the PQ score is an ideal way to accurately evaluate the hierarchical detection task.

Each submission has $3$ PQ scores for word, line, and paragraph respectively.
There are $3$ rankings for these $3$ sub-tasks respectively.
For the final ranking of the whole task, we compute the final score as a harmonic mean of the $3$ PQ scores (dubbed \textit{H-PQ}) and rank accordingly.

\subsubsection{Task 2: Word-Level End-to-End Text Detection and Recognition}
\begin{figure}
\centering
\includegraphics[width=1.0\textwidth]{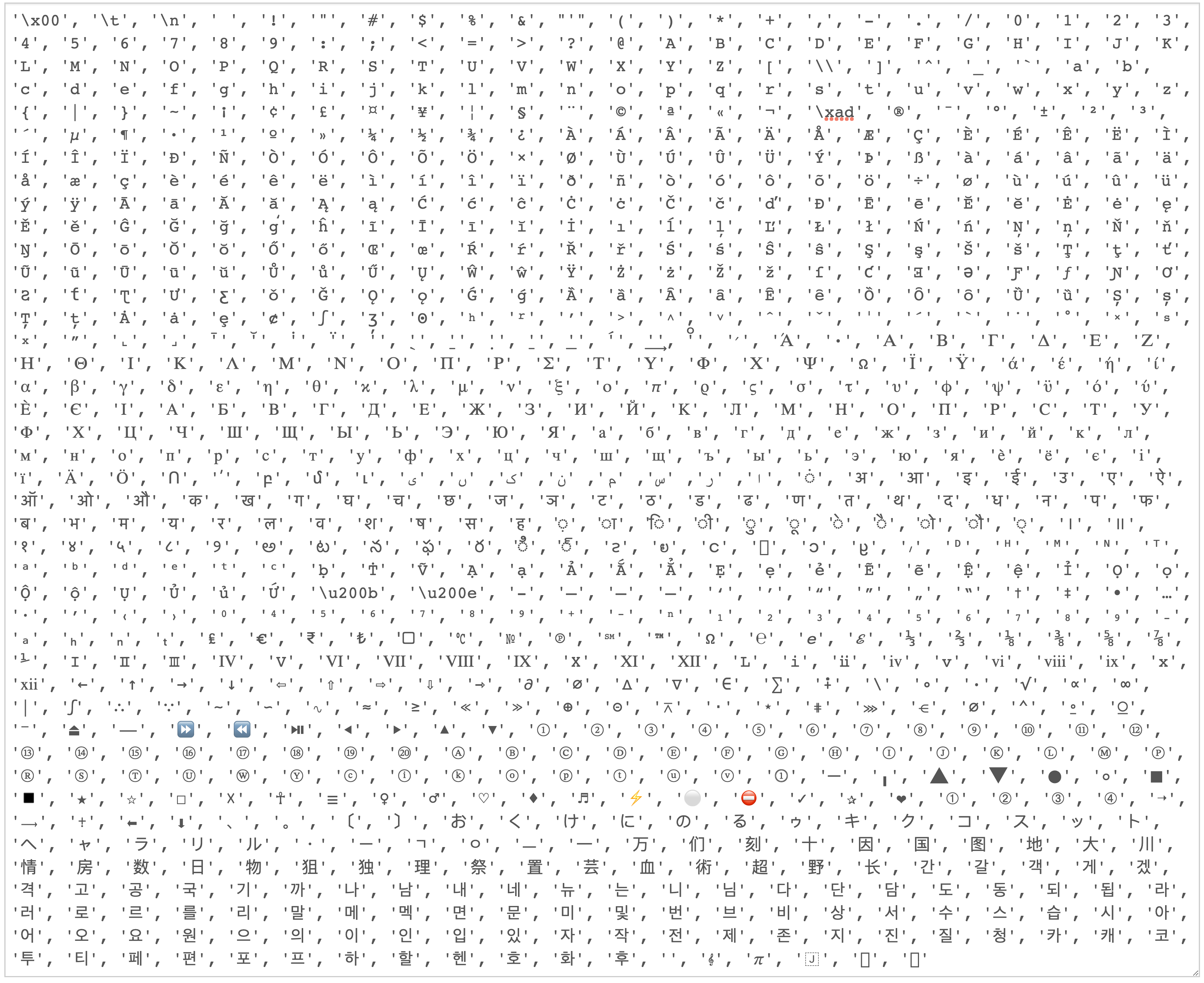}
\caption{
Character set in the training split.
}
\label{fig-char}
\end{figure}

For this task, images are provided and participants are expected to produce word-level text detection and recognition results, i.e. a set of word bounding polygons and transcriptions for each image.
Line and paragraph clustering is not required.
This is a challenging task, as the dataset has the most dense images, with more than $100$ words per image on average, $3$ times as many as the second dense dataset TextOCR \cite{singh_2021}. 
It also features a large number of recognizable characters. 
In the training set alone, there are more than $960$ different character classes, as shown in Fig. \ref{fig-char}, while most previous OCR benchmarks limit the tasks to recognize only digits and case-insensitive English characters.
These factors make this task challenging.

For evaluation, we use the F1 measure, which is a harmonic mean of word-level prediction and recall.
A word result is considered true positive if the IoU with ground-truth polygon is greater or equal to $0.5$ and the transcription is the same as the ground-truth.
The transcription comparison considers all characters and will be case-sensitive.
Note that, some words in the dataset are marked as illegible words.
Detection with high overlap with these words (IoU larger than $0.5$) will be removed in the evaluation process, and
ground-truths marked as illegible do not count as false negative even if they are not matched.

\subsection{Evaluation and Competition Website}
We host the competition on the widely recognized Robust Reading Competition (RRC) website\footnote{\url{https://rrc.cvc.uab.es/}} and set up our own competition page.
The RRC website has been the hub of scene text and document understanding research for a long time and hosted numerous prestigious competitions.
It provides easy-to-use infrastructure to set up competition, tasks, and carry out evaluation.
It also supports running the competition continuously, making it an ideal candidate.

\subsection{Competition Schedule}

We propose and execute the following competition schedule, in accordance with the conference timeline:

\begin{itemize}
  \item \textbf{January 2nd, 2023}: Start of the competition; submissions of results were enabled on the website.
  \item \textbf{April 1st, 2023}: Deadline for competition submissions.
  \item \textbf{April 15th, 2023}: Announcement of results.
\end{itemize}

\subsection{Other Competition Rules}
In addition to the aforementioned competition specifications, we also apply the following rules:

\begin{itemize}
  \item \textbf{Regarding the usage of other publicly available datasets}: HierText is the only allowed annotated OCR dataset. However, participants are also allowed to do self-labeling on other public OCR datasets as long as they don't use their ground-truth labels. In other words, they can use the images of other public datasets, but not their labels. They can also use non-OCR datasets, whether labeled or not, to pretrain their models. We believe they are important techniques that can benefit this field.
  \item \textbf{Usage of synthetic datasets} Synthetic data has been an important part of OCR recently \cite{synth1,synth2,synth3,synth4,synth5}. Participants can use any synthetic datasets, whether they are public or private, but are expected to reveal how they are synthesized and some basic statistics of the synthetic datasets if they are private.
  \item Participants should not use the validation split in training their models.
  \item Participants can make as many submissions as desired before the deadline, but we only archive the latest one submission of each participant in the final competition ranking. 
\end{itemize}

\subsection{Organizer Profiles}
Authors are all members of the OCR team at Google Research.
In addition to academic publications, authors have years of experience in building industrial OCR systems that are accurate and efficient for a diversity of image types and computation platforms.

\section{Competition Results}

\begin{figure}
\includegraphics[width=\textwidth]{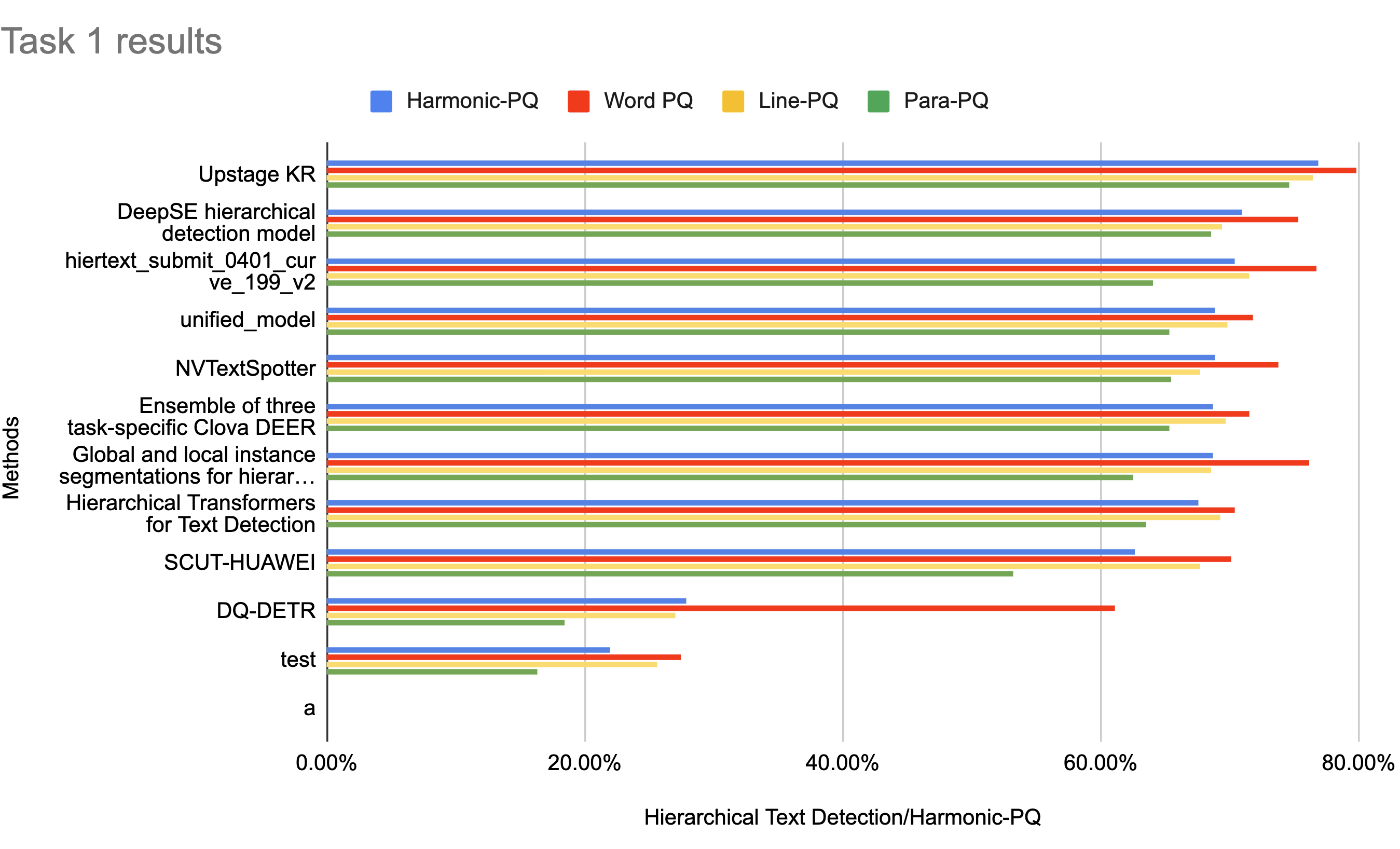}
\caption{
Figure for the results of task 1.
} \label{fig-task1}
\end{figure}

\begin{figure}
\includegraphics[width=\textwidth]{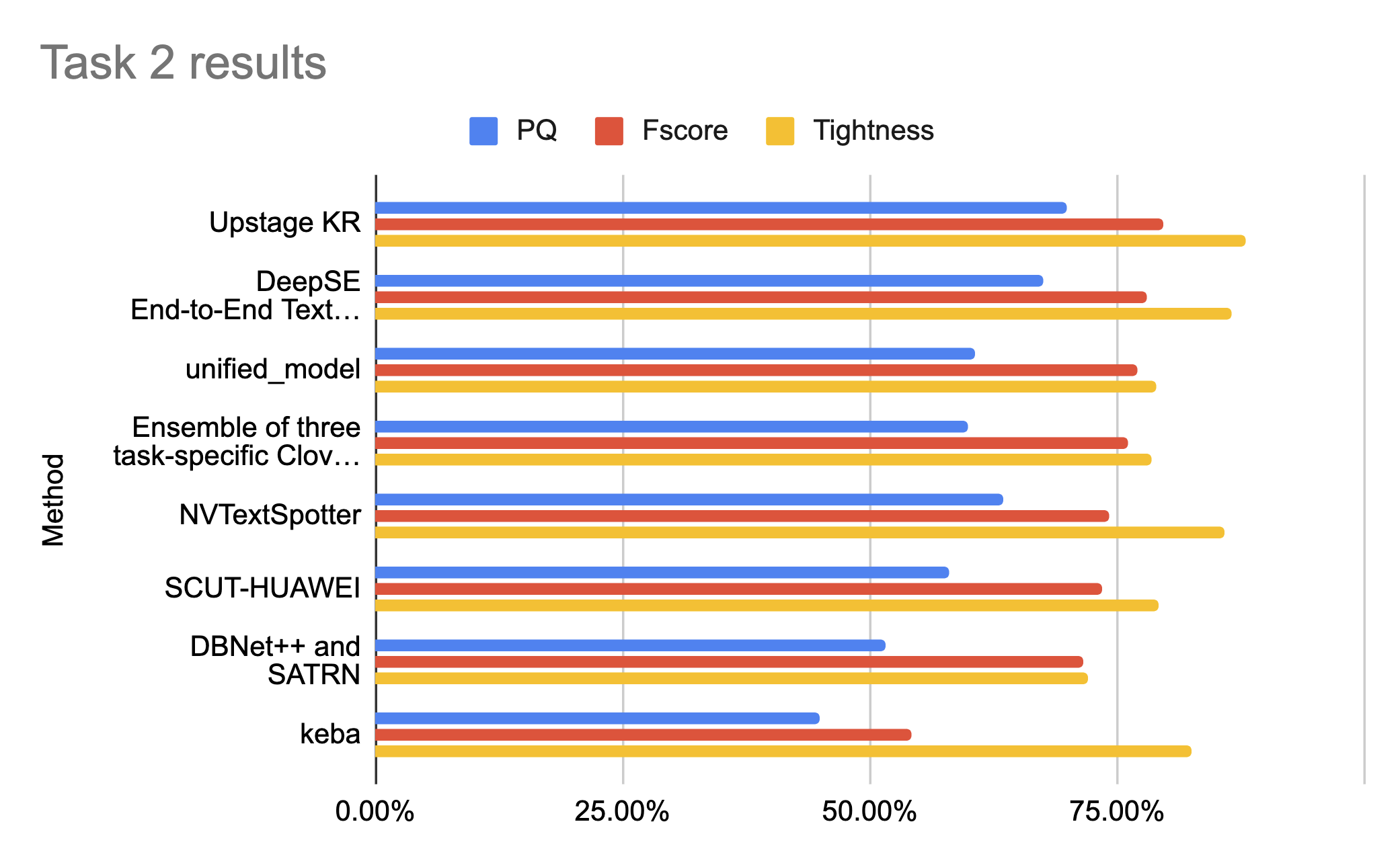}
\caption{
Figure for the results of task 2.
} \label{fig-task2}
\end{figure}

In total, the competition received $30$ submissions in Task 1 and $20$ submissions in Task 2. 
Note that, we encourage participants to submit multiple entries using different methods, for example, to understand the effect of applying different techniques such as pretraining and synthetic data.
To produce the final leaderboard in compliance with the ICDAR competition protocols, we only keep the latest $1$ submission from each participants.
The final deduplicated competition results are summarized in Tab. \ref{tab:task1} / Fig. \ref{fig-task1} and Tab. \ref{tab:task2} / Fig. \ref{fig-task2}.
In total, the competition received $11$ unique submissions in Task 1 and $7$ in Task 2.

% \usepackage{multirow}
% \usepackage{graphicx}
% \usepackage[normalem]{ulem}
\begin{table}[]
\resizebox{\columnwidth}{!}{%
\begin{tabular}{|l|l|l|r|rrrrr|rrrrr|rrrrr|}
\hline
{} & {} &  & \multicolumn{1}{l|}{{\textbf{\begin{tabular}[c]{@{}l@{}}Task 1\\ metric\end{tabular}}}} & \multicolumn{5}{c|}{{\textbf{Word}}} & \multicolumn{5}{c|}{{\textbf{Line}}} & \multicolumn{5}{c|}{{\textbf{Paragraph}}} \\ \cline{4-19} 
\multirow{-2}{*}{{User}} & \multirow{-2}{*}{{Method}} & \multirow{-2}{*}{\textbf{Rank}} & \multicolumn{1}{l|}{{\textbf{H-PQ}}} & \multicolumn{1}{l|}{{\textbf{PQ}}} & \multicolumn{1}{l|}{{\textbf{F}}} & \multicolumn{1}{l|}{{\textbf{P}}} & \multicolumn{1}{l|}{{\textbf{R}}} & \multicolumn{1}{l|}{{\textbf{T}}} & \multicolumn{1}{l|}{{\textbf{PQ}}} & \multicolumn{1}{l|}{{\textbf{F}}} & \multicolumn{1}{l|}{{\textbf{P}}} & \multicolumn{1}{l|}{{\textbf{R}}} & \multicolumn{1}{l|}{{\textbf{T}}} & \multicolumn{1}{l|}{{\textbf{PQ}}} & \multicolumn{1}{l|}{{\textbf{F}}} & \multicolumn{1}{l|}{{\textbf{P}}} & \multicolumn{1}{l|}{{\textbf{R}}} & \multicolumn{1}{l|}{{\textbf{T}}} \\ \hline
{YunSu Kim} & {Upstage KR} & 1 & {76.85} & \multicolumn{1}{r|}{{79.80}} & \multicolumn{1}{r|}{{91.88}} & \multicolumn{1}{r|}{{94.73}} & \multicolumn{1}{r|}{{89.20}} & {86.85} & \multicolumn{1}{r|}{{76.40}} & \multicolumn{1}{r|}{{88.34}} & \multicolumn{1}{r|}{{91.32}} & \multicolumn{1}{r|}{{85.56}} & {86.48} & \multicolumn{1}{r|}{{74.54}} & \multicolumn{1}{r|}{{86.15}} & \multicolumn{1}{r|}{{87.40}} & \multicolumn{1}{r|}{{84.94}} & {86.52} \\ \hline
{DeepSE x Upstage} & {\begin{tabular}[c]{@{}l@{}}DeepSE hierarchical\\ detection model\end{tabular}} & 2 & {70.96} & \multicolumn{1}{r|}{{75.30}} & \multicolumn{1}{r|}{{88.49}} & \multicolumn{1}{r|}{{93.50}} & \multicolumn{1}{r|}{{83.99}} & {85.10} & \multicolumn{1}{r|}{{69.43}} & \multicolumn{1}{r|}{{82.43}} & \multicolumn{1}{r|}{{82.65}} & \multicolumn{1}{r|}{{82.21}} & {84.23} & \multicolumn{1}{r|}{{68.51}} & \multicolumn{1}{r|}{{81.39}} & \multicolumn{1}{r|}{{81.69}} & \multicolumn{1}{r|}{{81.10}} & {84.17} \\ \hline
% {song} & {CLOVA DEER} & 3 & {70.66} & \multicolumn{1}{r|}{{74.20}} & \multicolumn{1}{r|}{{92.05}} & \multicolumn{1}{r|}{{93.86}} & \multicolumn{1}{r|}{{90.31}} & {80.61} & \multicolumn{1}{r|}{{71.40}} & \multicolumn{1}{r|}{{88.52}} & \multicolumn{1}{r|}{{91.56}} & \multicolumn{1}{r|}{{85.68}} & {80.66} & \multicolumn{1}{r|}{{66.79}} & \multicolumn{1}{r|}{{83.20}} & \multicolumn{1}{r|}{{82.92}} & \multicolumn{1}{r|}{{83.47}} & {80.28} \\ \hline
{zhm} & {\begin{tabular}[c]{@{}l@{}}hiertext\_submit\_0401\\ curve\_199\_v2\end{tabular}} & 3 & {70.31} & \multicolumn{1}{r|}{{76.71}} & \multicolumn{1}{r|}{{88.18}} & \multicolumn{1}{r|}{{92.71}} & \multicolumn{1}{r|}{{84.08}} & {86.99} & \multicolumn{1}{r|}{{71.43}} & \multicolumn{1}{r|}{{83.32}} & \multicolumn{1}{r|}{{89.32}} & \multicolumn{1}{r|}{{78.07}} & {85.73} & \multicolumn{1}{r|}{{63.97}} & \multicolumn{1}{r|}{{74.83}} & \multicolumn{1}{r|}{{81.25}} & \multicolumn{1}{r|}{{69.35}} & {85.48} \\ \hline
% {sukmin seo} & {\begin{tabular}[c]{@{}l@{}}D\_encoder decoder\\ segmentation\end{tabular}} & 5 & {68.92} & \multicolumn{1}{r|}{{71.82}} & \multicolumn{1}{r|}{{92.01}} & \multicolumn{1}{r|}{{93.75}} & \multicolumn{1}{r|}{{90.33}} & {78.06} & \multicolumn{1}{r|}{{69.73}} & \multicolumn{1}{r|}{{88.84}} & \multicolumn{1}{r|}{{91.28}} & \multicolumn{1}{r|}{{86.52}} & {78.49} & \multicolumn{1}{r|}{{65.51}} & \multicolumn{1}{r|}{{83.62}} & \multicolumn{1}{r|}{{84.32}} & \multicolumn{1}{r|}{{82.92}} & {78.34} \\ \hline
% {Taeho Kil} & {unified\_model} & 4 & {68.86} & \multicolumn{1}{r|}{{71.75}} & \multicolumn{1}{r|}{{91.95}} & \multicolumn{1}{r|}{{93.09}} & \multicolumn{1}{r|}{{90.83}} & {78.03} & \multicolumn{1}{r|}{{69.85}} & \multicolumn{1}{r|}{{89.00}} & \multicolumn{1}{r|}{{91.26}} & \multicolumn{1}{r|}{{86.86}} & {78.48} & \multicolumn{1}{r|}{{65.31}} & \multicolumn{1}{r|}{{83.50}} & \multicolumn{1}{r|}{{83.78}} & \multicolumn{1}{r|}{{83.22}} & {78.22} \\ \hline
{Mike Ranzinger} & {NVTextSpotter} & 4 & {68.82} & \multicolumn{1}{r|}{{73.69}} & \multicolumn{1}{r|}{{87.07}} & \multicolumn{1}{r|}{{95.10}} & \multicolumn{1}{r|}{{80.29}} & {84.63} & \multicolumn{1}{r|}{{67.76}} & \multicolumn{1}{r|}{{80.42}} & \multicolumn{1}{r|}{{93.87}} & \multicolumn{1}{r|}{{70.35}} & {84.25} & \multicolumn{1}{r|}{{65.51}} & \multicolumn{1}{r|}{{78.04}} & \multicolumn{1}{r|}{{81.82}} & \multicolumn{1}{r|}{{74.60}} & {83.94} \\ \hline
% {Kim Dae Hwan} & {SM Entertainment} & 8 & {68.81804} & \multicolumn{1}{r|}{{71.81}} & \multicolumn{1}{r|}{{91.92}} & \multicolumn{1}{r|}{{94.23}} & \multicolumn{1}{r|}{{89.72}} & {78.11} & \multicolumn{1}{r|}{{69.76}} & \multicolumn{1}{r|}{{88.81}} & \multicolumn{1}{r|}{{92.34}} & \multicolumn{1}{r|}{{85.54}} & {78.56} & \multicolumn{1}{r|}{{65.22}} & \multicolumn{1}{r|}{{83.20}} & \multicolumn{1}{r|}{{82.84}} & \multicolumn{1}{r|}{{83.57}} & {78.39} \\ \hline
% {ssm} & {ccn-transformer} & 9 & {68.80} & \multicolumn{1}{r|}{{71.58}} & \multicolumn{1}{r|}{{91.99}} & \multicolumn{1}{r|}{{93.66}} & \multicolumn{1}{r|}{{90.39}} & {77.81} & \multicolumn{1}{r|}{{69.79}} & \multicolumn{1}{r|}{{89.10}} & \multicolumn{1}{r|}{{91.73}} & \multicolumn{1}{r|}{{86.63}} & {78.32} & \multicolumn{1}{r|}{{65.32}} & \multicolumn{1}{r|}{{83.55}} & \multicolumn{1}{r|}{{83.17}} & \multicolumn{1}{r|}{{83.93}} & {78.19} \\ \hline
{ssm} & {\begin{tabular}[c]{@{}l@{}}Ensemble of three \\ task-specific \\ Clova DEER detection\end{tabular}} & 5 & {68.72} & \multicolumn{1}{r|}{{71.54}} & \multicolumn{1}{r|}{{92.03}} & \multicolumn{1}{r|}{{93.82}} & \multicolumn{1}{r|}{{90.31}} & {77.74} & \multicolumn{1}{r|}{{69.64}} & \multicolumn{1}{r|}{{89.04}} & \multicolumn{1}{r|}{{91.75}} & \multicolumn{1}{r|}{{86.49}} & {78.21} & \multicolumn{1}{r|}{{65.29}} & \multicolumn{1}{r|}{{83.70}} & \multicolumn{1}{r|}{{84.17}} & \multicolumn{1}{r|}{{83.23}} & {78.01} \\ \hline
{xswl} & {\begin{tabular}[c]{@{}l@{}}Global and local instance\\ segmentations for \\ hierarchical text detection\end{tabular}} & 6 & {68.62} & \multicolumn{1}{r|}{{76.16}} & \multicolumn{1}{r|}{{90.72}} & \multicolumn{1}{r|}{{93.45}} & \multicolumn{1}{r|}{{88.16}} & {83.95} & \multicolumn{1}{r|}{{68.50}} & \multicolumn{1}{r|}{{82.22}} & \multicolumn{1}{r|}{{80.24}} & \multicolumn{1}{r|}{{84.31}} & {83.31} & \multicolumn{1}{r|}{{62.55}} & \multicolumn{1}{r|}{{75.11}} & \multicolumn{1}{r|}{{74.00}} & \multicolumn{1}{r|}{{76.25}} & {83.28} \\ \hline
% {Kim Dong Hyun} & {The tight prediction} & 12 & {68.48} & \multicolumn{1}{r|}{{71.41}} & \multicolumn{1}{r|}{{91.93}} & \multicolumn{1}{r|}{{93.59}} & \multicolumn{1}{r|}{{90.33}} & {77.68} & \multicolumn{1}{r|}{{69.49}} & \multicolumn{1}{r|}{{88.93}} & \multicolumn{1}{r|}{{91.78}} & \multicolumn{1}{r|}{{86.25}} & {78.14} & \multicolumn{1}{r|}{{64.88}} & \multicolumn{1}{r|}{{83.25}} & \multicolumn{1}{r|}{{83.69}} & \multicolumn{1}{r|}{{82.81}} & {77.94} \\ \hline
{Asaf Gendler} & {\begin{tabular}[c]{@{}l@{}}Hierarchical Transformers \\ for Text Detection\end{tabular}} & 7 & {67.59} & \multicolumn{1}{r|}{{70.44}} & \multicolumn{1}{r|}{{86.09}} & \multicolumn{1}{r|}{{88.47}} & \multicolumn{1}{r|}{{83.83}} & {81.82} & \multicolumn{1}{r|}{{69.30}} & \multicolumn{1}{r|}{{85.23}} & \multicolumn{1}{r|}{{87.83}} & \multicolumn{1}{r|}{{82.78}} & {81.31} & \multicolumn{1}{r|}{{63.46}} & \multicolumn{1}{r|}{{78.40}} & \multicolumn{1}{r|}{{77.84}} & \multicolumn{1}{r|}{{78.97}} & {80.94} \\ \hline
% {Wang Dong Hyun} & {Beautiful SKY} & 14 & {68.24} & \multicolumn{1}{r|}{{70.91}} & \multicolumn{1}{r|}{{91.57}} & \multicolumn{1}{r|}{{93.73}} & \multicolumn{1}{r|}{{89.50}} & {77.44} & \multicolumn{1}{r|}{{69.05}} & \multicolumn{1}{r|}{{88.52}} & \multicolumn{1}{r|}{{92.03}} & \multicolumn{1}{r|}{{85.27}} & {78.01} & \multicolumn{1}{r|}{{65.04}} & \multicolumn{1}{r|}{{83.53}} & \multicolumn{1}{r|}{{83.43}} & \multicolumn{1}{r|}{{83.63}} & {77.87} \\ \hline
{JiangQing} & {SCUT-HUAWEI} & 8 & {62.68} & \multicolumn{1}{r|}{{70.08}} & \multicolumn{1}{r|}{{89.58}} & \multicolumn{1}{r|}{{89.79}} & \multicolumn{1}{r|}{{89.37}} & {78.23} & \multicolumn{1}{r|}{{67.70}} & \multicolumn{1}{r|}{{86.20}} & \multicolumn{1}{r|}{{90.46}} & \multicolumn{1}{r|}{{82.33}} & {78.53} & \multicolumn{1}{r|}{{53.14}} & \multicolumn{1}{r|}{{69.06}} & \multicolumn{1}{r|}{{74.03}} & \multicolumn{1}{r|}{{64.72}} & {76.96} \\ \hline
{Jiawei Wang} & {DQ-DETR} & 9 & {27.81} & \multicolumn{1}{r|}{{61.01}} & \multicolumn{1}{r|}{{77.27}} & \multicolumn{1}{r|}{{80.64}} & \multicolumn{1}{r|}{{74.17}} & {78.96} & \multicolumn{1}{r|}{{26.96}} & \multicolumn{1}{r|}{{35.91}} & \multicolumn{1}{r|}{{26.81}} & \multicolumn{1}{r|}{{54.39}} & {75.07} & \multicolumn{1}{r|}{{18.38}} & \multicolumn{1}{r|}{{24.72}} & \multicolumn{1}{r|}{{15.99}} & \multicolumn{1}{r|}{{54.41}} & {74.36} \\ \hline
{ZiqianShao} & {test} & 10 & {21.94} & \multicolumn{1}{r|}{{27.45}} & \multicolumn{1}{r|}{{41.75}} & \multicolumn{1}{r|}{{51.82}} & \multicolumn{1}{r|}{{34.95}} & {65.76} & \multicolumn{1}{r|}{{25.61}} & \multicolumn{1}{r|}{{39.04}} & \multicolumn{1}{r|}{{51.50}} & \multicolumn{1}{r|}{{31.43}} & {65.59} & \multicolumn{1}{r|}{{16.32}} & \multicolumn{1}{r|}{{24.52}} & \multicolumn{1}{r|}{{35.61}} & \multicolumn{1}{r|}{{18.70}} & {66.57} \\ \hline
{Yichuan Cheng} & {a} & 11 & {0.00} & \multicolumn{1}{r|}{{0.00}} & \multicolumn{1}{r|}{{0.00}} & \multicolumn{1}{r|}{{0.24}} & \multicolumn{1}{r|}{{0.00}} & {53.62} & \multicolumn{1}{r|}{{0.01}} & \multicolumn{1}{r|}{{0.01}} & \multicolumn{1}{r|}{{0.25}} & \multicolumn{1}{r|}{{0.01}} & {51.29} & \multicolumn{1}{r|}{{0.01}} & \multicolumn{1}{r|}{{0.02}} & \multicolumn{1}{r|}{{0.21}} & \multicolumn{1}{r|}{{0.01}} & {50.89} \\ \hline
\end{tabular}%
}
\caption{
Results for Task 1.
F/P/R/T/PQ stand for \textit{F1-score}, \textit{Precision}, \textit{Recall}, \textit{Tightness}, and \textit{Panoptic Quality} respectively.
The submissions are ranked by the \textit{H-PQ} score.
H-PQ can be interpreted as \textit{Hierarchical-PQ} or \textit{Harmonic-PQ}. 
H-PQ is calculated as the harmonic means of the PQ scores of the $3$ hierarchies: word, line, and paragraph.
It represents the comprehensive ability of a method to detect the text hierarchy in image.
We omit the $\%$ for all these numbers for simplicity.
}
\label{tab:task1}
\end{table}

% \usepackage{multirow}
% \usepackage{graphicx}
% \usepackage[normalem]{ulem}
\begin{table}[]
\resizebox{\columnwidth}{!}{%
\begin{tabular}{|l|l|l|rrrrr|}
\hline
{} & {} &  & \multicolumn{5}{c|}{{ \textbf{Word}}} \\ \cline{4-8} 
\multirow{-2}{*}{{\textbf{User}}} & \multirow{-2}{*}{{\textbf{Method}}} & \multirow{-2}{*}{\textbf{Rank}} & \multicolumn{1}{l|}{{ \textbf{PQ}}} & \multicolumn{1}{l|}{{ \textbf{F}}} & \multicolumn{1}{l|}{{ \textbf{P}}} & \multicolumn{1}{l|}{{ \textbf{R}}} & \multicolumn{1}{l|}{{ \textbf{T}}} \\ \hline
{ YunSu Kim} & {{Upstage KR}} & 1 & \multicolumn{1}{r|}{{ 70.00}} & \multicolumn{1}{r|}{{ 79.58}} & \multicolumn{1}{r|}{{ 82.05}} & \multicolumn{1}{r|}{{ 77.25}} & { 87.97} \\ \hline
{ DeepSE x Upstage} & {{DeepSE End-to-End Text Detection and Recognition Model}} & 2 & \multicolumn{1}{r|}{{ 67.46}} & \multicolumn{1}{r|}{{ 77.93}} & \multicolumn{1}{r|}{{ 88.05}} & \multicolumn{1}{r|}{{ 69.89}} & { 86.57} \\ \hline
% { Kim Dong Hyun} & {{The tight prediction}} & 3 & \multicolumn{1}{r|}{{ 60.53}} & \multicolumn{1}{r|}{{ 77.09}} & \multicolumn{1}{r|}{{ 78.48}} & \multicolumn{1}{r|}{{ 75.75}} & { 78.52} \\ \hline
% { Taeho Kil} & {{unified\_model}} & 3 & \multicolumn{1}{r|}{{ 60.70}} & \multicolumn{1}{r|}{{ 76.95}} & \multicolumn{1}{r|}{{ 77.91}} & \multicolumn{1}{r|}{{ 76.02}} & { 78.89} \\ \hline
{ ssm} & {{Ensemble of three task-specific Clova DEER}} & 3 & \multicolumn{1}{r|}{{ 59.84}} & \multicolumn{1}{r|}{{ 76.15}} & \multicolumn{1}{r|}{{ 77.63}} & \multicolumn{1}{r|}{{ 74.73}} & { 78.59} \\ \hline
% { ssm} & {{cnn-transformer}} & 6 & \multicolumn{1}{r|}{{ 59.87}} & \multicolumn{1}{r|}{{ 76.10}} & \multicolumn{1}{r|}{{ 77.48}} & \multicolumn{1}{r|}{{ 74.77}} & { 78.67} \\ \hline
% { Kim Dae Hwan} & {{SM Entertainment}} & 7 & \multicolumn{1}{r|}{{ 59.84}} & \multicolumn{1}{r|}{{ 75.77}} & \multicolumn{1}{r|}{{ 77.68}} & \multicolumn{1}{r|}{{ 73.96}} & { 78.98} \\ \hline
% { Wang Dong Hyun} & {{Beautiful SKY}} & 8 & \multicolumn{1}{r|}{{ 59.13}} & \multicolumn{1}{r|}{{ 75.51}} & \multicolumn{1}{r|}{{ 77.30}} & \multicolumn{1}{r|}{{ 73.80}} & { 78.30} \\ \hline
% { song} & {{CLOVA DEER}} & 9 & \multicolumn{1}{r|}{{ 60.36}} & \multicolumn{1}{r|}{{ 74.21}} & \multicolumn{1}{r|}{{ 75.27}} & \multicolumn{1}{r|}{{ 73.19}} & { 81.33} \\ \hline
{ Mike Ranzinger} & {{NVTextSpotter}} & 4 & \multicolumn{1}{r|}{{ 63.57}} & \multicolumn{1}{r|}{{ 74.10}} & \multicolumn{1}{r|}{{ 80.94}} & \multicolumn{1}{r|}{{ 68.34}} & { 85.78} \\ \hline
{ JiangQing} & {{SCUT-HUAWEI}} & 5 & \multicolumn{1}{r|}{{ 58.12}} & \multicolumn{1}{r|}{{ 73.41}} & \multicolumn{1}{r|}{{ 74.38}} & \multicolumn{1}{r|}{{ 72.46}} & { 79.17} \\ \hline
{ kuli.cyd} & {{DBNet++ and SATRN}} & 6 & \multicolumn{1}{r|}{{ 51.62}} & \multicolumn{1}{r|}{{ 71.64}} & \multicolumn{1}{r|}{{ 82.76}} & \multicolumn{1}{r|}{{ 63.15}} & { 72.06} \\ \hline
{ LGS} & {{keba}} & 7 & \multicolumn{1}{r|}{{ 44.87}} & \multicolumn{1}{r|}{{ 54.30}} & \multicolumn{1}{r|}{{ 68.37}} & \multicolumn{1}{r|}{{ 45.03}} & { 82.64} \\ \hline
\end{tabular}%
}
\caption{
Results for Task 2.
F/P/R/T/PQ stand for \textit{F1-score}, \textit{Precision}, \textit{Recall}, \textit{Tightness}, and \textit{Panoptic Quality} respectively.
The submissions are ranked by the F1 score.
We omit the $\%$ for all these numbers for simplicity.
}
\label{tab:task2}
\end{table}

\subsection{Submission Validation}

In the final leaderboard, each participant is only allowed to have one submission.
We validate each submission and examine the number of submissions from each team. 
If a team has more than one submission, we keep the latest one and remove the rest from the leaderboard.
Note that these removed submissions will remain on the RRC portal for reference, since they also provide important aspects into this research field. 
We adopt the following rules to determine the authorship of each submission:

\begin{itemize}
  \item \textbf{user\_id}: If two submissions have the same user\_id field, it means they are submitted by the same RRC user account and thus should be from the same team.
  \item \textbf{method description}: Participants are asked to provide descriptive information of their submissions, including authors, method details, etc. If two submissions have strictly almost identical author list and method description, we consider them to be from the same team.
\end{itemize}

\subsection{Task 1 Methodology}
Task 1 in our competition, i.e. Hierarchical Text Detection, is a novel task in the research field. 
There are no existing methods that participants can refer to. 
Even the previous work Unified Detector \cite{long_2022} can only produce line and paragraph outputs but no word-level results.
Among the $8$ submissions in Task 1 which have disclosed their methods, we observed that $5$ of them develop `\textit{multi-head plus postprocessing}' systems.
These methods treat words, lines, and paragraphs as generic objects, and train detection or segmentation models to localize these three levels of text entities in parallel with separate prediction branches for each level. 
In the post-processing, they use IoU-based rules to build the hierarchy in the post-processing step, i.e. assigning words to lines and lines to paragraphs.
The most of the top ranking solutions belong to this type of methods.
One submission (from the SCUT-HUAWEI team) adopts a cascade pipeline, by first detecting words  and then applying LayoutLMv3 \cite{huang_2022} to cluster words into lines and paragraphs. 
The \textit{Hierarchical Transformers for Text Detection} method develops a unified detector similar to \cite{long_2022} for line detection and paragraph grouping and also a line-to-word detection model that produces bounding boxes for words.
Here we briefly introduce the top 2 methods in this task:

\noindent \textbf{Upstage KR team} ranks 1st place in Task 1, achieving an H-PQ metric of $76.85\%$. It beats the second place by almost $6\%$ in the H-PQ metric. They implemented a two-step approach to address hierarchical text detection. First, they performed multi-class semantic segmentation where classes were word, line, and paragraph regions. Then, they used the predicted probability map to extract and organize these entities hierarchically. Specifically, an ensemble of UNets with ImageNet-pretrained EfficientNetB7\cite{efficient} / MitB4 \cite{segformer} backbones was utilized to extract class masks.
Connected components were identified in the predicted mask to separate words from each other, same for lines and paragraphs. Then, a word was assigned as a child of a line if the line had the highest IoU with the word compared to all other lines. This process was similarly applied to lines and paragraphs. For training, they eroded target entities and dilated predicted entities. 
Also, they ensured that target entities maintained a gap between them. They used symmetric Lovasz loss \cite{lovasz} and pre-trained their models on the SynthText dataset \cite{synth4}.

\noindent \textbf{DeepSE X Upstage HK team} ranks 2nd in the leaderboard. They fundamentally used DBNet \cite{dbnet} as the scene text detector, and leveraged the oCLIP \cite{oclip} pretrained Swin Transformer-Base \cite{swin} model as the backbone to make direct predictions at three different levels. Following DBNet, they employed Balanced Cross-Entropy for binary map and L1 loss for threshold map. The authors also further fine-tuned the model with lovasz loss \cite{lovasz} for finer localization.

\subsection{Task 2 Methodology}
Task 2, i.e. Word-Level End-to-End Text Detection and Recognition, is a more widely studied task.
Recent research \cite{qin_2019,solo} focuses on building end-to-end trainable OCR models, as opposed to separately trained detection and recognition models.
It's widely believed that end-to-end models enjoy shared feature extraction which leads to better accuracy.
However, the results of our competition say otherwise.
The top 2 methods by the \textbf{Upstage KR team} and \textbf{DeepSE End-to-End Text Detection and Recognition Model team} are all separately trained models.
There are two end-to-end submissions.
The \textbf{unified\_model team} applies a deformable attention decoder based text recognizer and ranks 3th place.
% The \textbf{CLOVA DEER team} bases their solution on the DEER end-to-end OCR model \cite{deer} and ranks 9th place.
Here we briefly introduce the top 2 methods in this task:

\noindent \textbf{Upstage KR team} uses the same task 1 method for detecting words.
For word-level text recognition, they use the ParSeq \cite{parseq} model but replace the visual feature extractor with SwinV2 \cite{swin2}.
The text recognizer is pretrained with synthetic data before fine-tuning it on the HierText dataset.
They use an in-house synthetic data generator derived from the open source SynthTiger \cite{synth5} to generate word images using English and Korean corpus. Notably, they generate 5M English/Korean word images with vertical layout, in addition to 10M English / Korean word images with horizontal layout.
For the final submission, they use an ensemble of three text recognizers for strong and stable performance.

\noindent \textbf{DeepSE End-to-End Text Detection and Recognition Model team} also uses the ParSeq \cite{parseq} model as their recognizer.
They point out that, in order to make the data domain consistent between the training and inference stages, they run their detector on training data, and then crop words using detected boxes.
This step is important int adapting the training domain to the inference domain. This trick essentially improves their model's performance.

\section{Discussion}
\begin{figure}
\includegraphics[width=\textwidth]{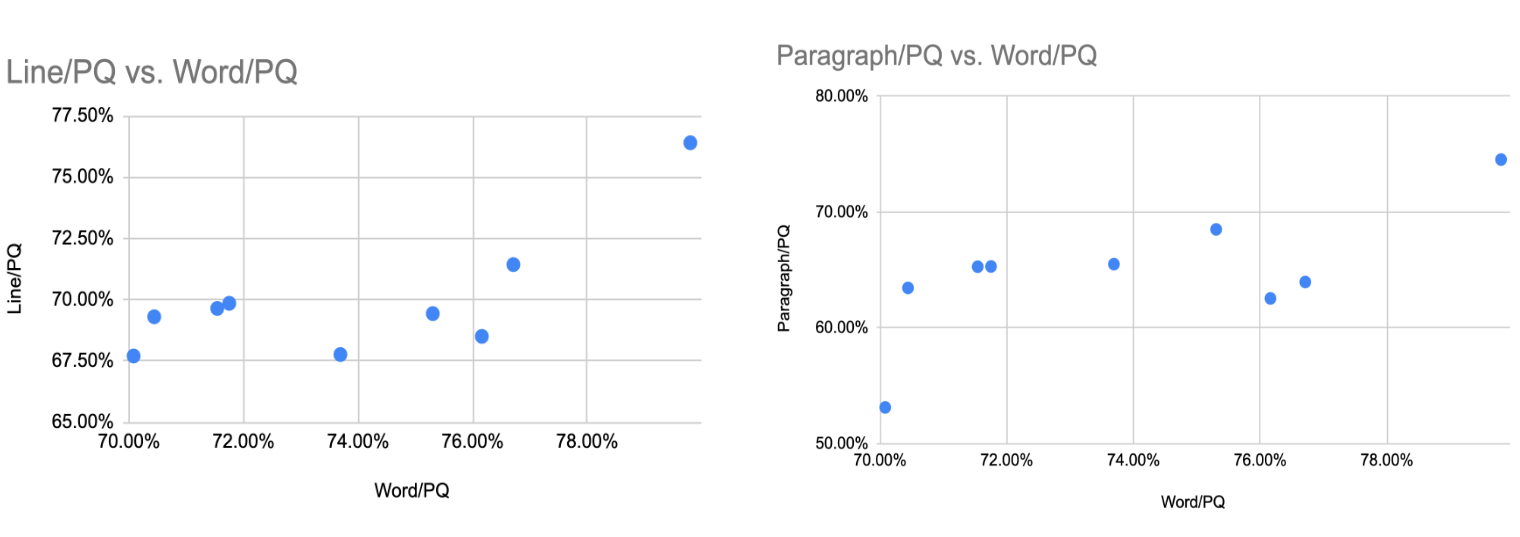}
\caption{
Correlation between text levels.
Each dot is a submission in the Task 1.
\textbf{Left}: Correlation between word PQ and line PQ.
\textbf{Right}: Correlation between word PQ and paragraph PQ.
} \label{fig-corr}
\end{figure}

In the Hierarchical Text Detection task, the original Unified Detector \cite{long_2022} can only achieve PQ scores of $48.21\%$, $62.23\%$, $53.60\%$ on the words, lines, and paragraphs respectively.
The H-PQ score for Unified Detector is only $54.08\%$, ranking at 10th place if put in the competition leaderboard.
The winning solution exceeds Unified Detector by more than $20\%$.
These submissions greatly push the envelope of state-of-the-art Hierarchical Text Detection method.
However, current methods are still not satisfactory. 
As shown in Fig. \ref{fig-task1}, we can easily notice that for all methods, word PQ scores are much higher than line PQ scores, and line PQ scores are again much higher than paragraph PQ scores.
It indicates that, line and paragraph level detections are still more difficult than word detection.
Additionally, Fig. \ref{fig-corr} shows that layout analysis performance is only marginally correlated with word detection performance, especially when outliers are ignored. We believe there's still hidden challenges and chances for improvement in layout analysis.
Furthermore, winning solutions in our competition rely on postprocessing which can be potentially complicated and error-prone.
It's also important to improve end-to-end methods.

The task 2 of our challenge is a standard yet unique end-to-end detection and recognition task.
While it inherits the basic setting of an end-to-end task, it is based on a diversity of images which has high word density, and it has an unlimited character set. 
For this task, we see most of the submissions are two-stage methods, where the detection and recognition models are trained separately, and there's no feature sharing.
These two-stage methods achieve much better performances than end-to-end submissions.
This contrasts with the trend in research paper that favors end-to-end trainable approaches with feature sharing between the two stage.
Therefore, we believe the HierText dataset can be a very useful benchmark in end-to-end OCR research.
Another interesting observation for Task 2 is that, while most submissions achieve a tightness score of around $80\%$, the correlation between tightness scores and F1 scores and very low, with a correlation coefficient of $0.06$.
It could indicate that recognition is less sensitive to the accuracy of bounding boxes after it surpasses some threshold.
This would mean that the mainstream training objective of maximizing bounding box IoU might not be the optimal target.
For example, a slightly oversized bounding box is better than a small one which might miss some characters.
With that said, a precise bounding box is still useful itself, which indicates the localization.
Another potential reason is that bounding box annotation is not always accurate -- it's always oversized because text are not strictly rectangular.

\section{Conclusion}
This paper summarizes the organization and results of ICDAR 2023 Competition on Hierarchical Text Detection and Recognition.
We share details of competition motivation, dataset collection, competition organization, and result analysis.
In total, we have $18$ valid and unique competition entries, showing great interest from both research communities and industries.
We keep the competition submission site open to promote research into this field.
We also plan to extend and improve this competition, for example, adding multilingual data.


\end{document}